%% file: main.tex
\documentclass[letterpaper, 10 pt, conference]{ieee_format/ieeeconf}

\IEEEoverridecommandlockouts                              

\usepackage[noadjust]{cite}
\usepackage{threeparttable}
\usepackage{url}

\usepackage{xcolor}
\usepackage{kotex}

\usepackage{graphicx}
\usepackage{amssymb} 
\usepackage{amsmath}

\usepackage{siunitx}
\usepackage{float}
\usepackage{dsfont}
\usepackage{multicol}
\usepackage{bm}
\usepackage{booktabs}
\usepackage{tabularx}
\let\labelindent\relax
\usepackage{enumitem}

\usepackage{multirow}

\usepackage{caption}
\title{\LARGE \bf
    Learning-based Adaptive Control of Quadruped Robots 
    
    for Active Stabilization on Moving Platforms
}

\author{Minsung Yoon\textsuperscript{1}, Heechan Shin\textsuperscript{1}, Jeil Jeong\textsuperscript{2}, and Sung-Eui Yoon\textsuperscript{1}\textsuperscript{\textdagger}
\thanks{
    \textsuperscript{1}M. Yoon, H. Shin, and S. Yoon are with the School of Computing at Korea Advanced Institute of Science and Technology (KAIST), Daejeon, 34141, Republic of Korea.
    \textsuperscript{2}J. Jeong is with the Robotics Program at the same institute, KAIST. 
    {\textsuperscript{\textdagger}}S. Yoon is a corresponding author; {\tt\small sungeui@kaist.edu}.
    }}

\begin{document}
    \maketitle
    \thispagestyle{empty}
    \pagestyle{empty}

    \input{Paper/0_abs}
    \input{Paper/1_intro}
    \input{Paper/2_related_work}

\input{Paper/3_method}
    \input{Paper/4_experiment}
    \input{Paper/6_conclusion}

    {
        \small
        \bibliographystyle{ieee_format/ieee}
        \bibliography{./ref}
    }

\end{document}

%% file: Paper/0_abs.tex
\begin{abstract}
A quadruped robot faces balancing challenges on a six-degrees-of-freedom moving platform, like subways, buses, airplanes, and yachts, due to independent platform motions and resultant diverse inertia forces on the robot. To alleviate these challenges, we present the Learning-based Active Stabilization on Moving Platforms (\textit{LAS-MP}), featuring a self-balancing policy and system state estimators. The policy adaptively adjusts the robot's posture in response to the platform's motion. The estimators infer robot and platform states based on proprioceptive sensor data. For a systematic training scheme across various platform motions, we introduce platform trajectory generation and scheduling methods. Our evaluation demonstrates superior balancing performance across multiple metrics compared to three baselines. Furthermore, we conduct a detailed analysis of the \textit{LAS-MP}, including ablation studies and evaluation of the estimators, to validate the effectiveness of each component.
\end{abstract}

%% file: Paper/1_intro.tex
\section{Introduction}

Quadruped robots have become essential in various real-world applications, such as rescue missions and surveillance, thanks to their adeptness at navigating diverse terrains~\cite{bellicoso2018advances, delmerico2019current}. 
A crucial aspect of their deployment is the ability to maintain stability and balance while traversing different types of terrain, including rough and slippery surfaces.
To do that, significant progress has been made in various ways, including Reinforcement Learning (RL)-based control strategies, terrain property estimation, foot-terrain interaction analysis, and adaptive mechanical foot designs (refer to Sec.~\ref{sec:2-A}).

On the other hand, as quadruped robots are increasingly deployed in inhabited environments, they have opportunities to utilize transportation platforms like subways, buses, yachts, airplanes, and conveyors for efficient navigation.
However, robots in non-inertial reference frames (e.g., moving platforms) encounter distinct balancing challenges compared to those on terrains which are inertial systems. 
As an example, we can first imagine fictitious inertial forces experienced by passengers in accelerating or turning vehicles. 
In addition, the absence of prior knowledge about platform movements further complicates prompt reaction to the forces to maintain balance.
These challenges highlight the need for advanced control strategies to swiftly adapt to continuously changing platform motions to ensure a balance of the robot.

\begin{figure}[t]
    \vspace{0.2cm}
    \centering 
    \includegraphics[width=0.9\columnwidth]{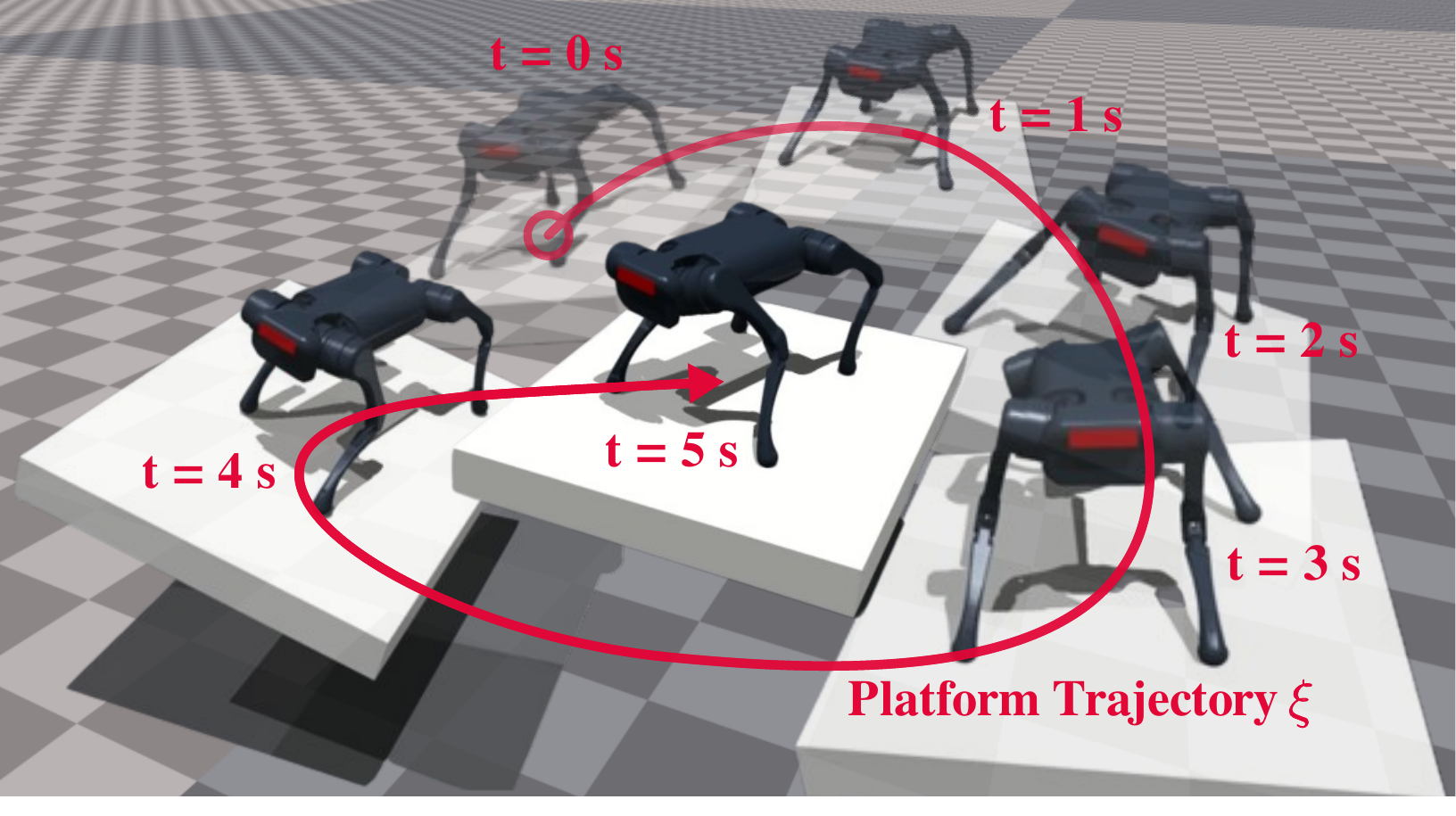}
	\caption{
        An illustration demonstrating a balancing process of a quadruped robot on a moving platform exhibiting six-degrees-of-freedom motions.
	}
	\label{fig:fig1}
	\vspace{-0.65cm}
\end{figure}

Thus, our goal is to ensure that quadruped robots maintain balance and avoid falling off moving platforms even with limited space, as illustrated in Fig.~\ref{fig:fig1}.
Achieving this requires timely and adaptive postural adjustments based on an understanding of platform movements. 
To the best of our knowledge, we believe this work is the first effort to address the challenges associated with six-degrees-of-freedom (DoF) and unknown platform motions in three-dimensional space across the fields of quadrupedal locomotion and self-balancing research (refer to Sec.~\ref{sec:2-B},~\ref{sec:2-C}).
As the number of dimensions of the platform's motions increases, the robot encounters a broader array of inertial forces, complicating the balancing task, as further detailed in Sec.~\ref{sec:3-A}.

\textbf{Main Contributions}.  
We introduce the Learning-based Active Stabilization method on Moving Platforms (\textit{LAS-MP}), designed for adaptive posture adjustment in response to 6-DoF platform motions. 
This method features a self-balancing policy and system state estimators developed via RL with system identification (refer to Sec.~\ref{sec:2-D}). 
The estimators infer system states related to the robot and platform (e.g., velocities) alongside the robot's intrinsic properties (e.g., friction), utilizing historical proprioceptive sensor data.
These inferred states improve the situational awareness of the policy in non-inertial frames, facilitating adaptive balancing on moving platforms. 
In addition, we augment the policy with an alignment command, derived from the robot and platform states via feature engineering, which provides explicit velocity guidance to the policy so that it converges better solutions.

To train the policy in environments consisting of diverse platform motions, we introduce platform trajectory generators based on basis-splines (B-splines)~\cite{knott1999interpolating}. 
Furthermore, we present a curriculum and scheduling process to progressively escalate task complexity for successful policy learning.

In experiments, we show that the \textit{LAS-MP} enhances the balancing performance across various platform movements, outperforming three other baselines.
We also indicate that baseline systems, which are competent in balancing on rough and stationary terrains, struggle on moving platforms. 
In addition, we carry out ablation studies to elucidate the impact of explicitly estimating the robot and platform states and the effectiveness of the alignment command. 

%% file: Paper/2_related_work.tex
\section{Related Work}

\subsection{Quadrupedal Locomotion on Various Terrains}
\label{sec:2-A}
Despite advancements in quadrupedal locomotion on diverse terrains,
traditional model-based methods encounter challenges in model accuracy, uncertainty, and adaptability~\cite{fankhauser2018robust, mastalli2022agile, lu2023whole}.
Recent deep Reinforcement Learning (RL) offers promising solutions to these challenges~\cite{9779429, nahrendra2023dreamwaq, kumar2021rma, agarwal2023legged, lee2020learning, 9981198, cheng2023parkour, rudin2022learning}.
However, RL methods still face limitations in handling various physical properties of the terrains, such as friction and restitution, when relying solely on geometrical sensor data. 
Thus, researchers have devised specialized strategies tailored to elastic~\cite{kim2021quadruped}, soft~\cite{8957061, yao2023predict, Suyoung2023deformter}, and slippery~\cite{8772165, Focchi2018slip, s22082967} terrains, including foot-terrain interaction modeling and abnormal state detection. 
In addition, some RL studies~\cite{rudin2022learning, cheng2023parkour} simulate virtual external forces on robots to boost the robustness of the locomotion policy. 
However, these stability challenges only partially resemble the range of perturbations from moving platforms.
Thus, we focus on ensuring the robot's safety on the platform in the face of active platform movements.

\subsection{Quadrupedal Locomotion on Moving Platforms}
\label{sec:2-B}
Focusing on operations on moving platforms, recent locomotion research advances traditional model-based controllers and gait planners by integrating platform motions into the analytical models~\cite{iqbal2020provably, iqbal2021extended}. 
However, their applicability to diverse platform motions is limited by several assumptions, such as known platform motions with negligible horizontal acceleration, constant body height, negligible angular momentum, and fixed walking cycles with three contact points.
These studies explored 2-DoF platform movements featuring sinusoidal pitching and vertical motions, emulating the vessel movements in regular sea waves.
In contrast, our research investigates more complex scenarios with 6-DoF platform motions without imposing constraints on the robot and platform's motions.
Furthermore, our method operates without prior knowledge of the underlying platform motions, estimating them through state estimators from historical proprioceptive sensor data to adapt to the platform motions.

\subsection{Quadrupedal Self-balancing on Moving Platforms} 
\label{sec:2-C}
Rather than the locomotion, other researchers have concentrated on enhancing the postural stabilization of the robot on tilting platforms~\cite{li2022posture, grimminger2020open, lee2021reinforcement, sun2021adaptive}. 
A recent RL-based approach formed an optimal self-balancing policy using table-based RL, integrating kinematic equations but neglecting physical factors like gravity and contact dynamics~\cite{lee2021reinforcement}. 
A subsequent study solved a continuous optimization problem, employing a parameterized stochastic policy with a physics simulation~\cite{sun2021adaptive}. 
While these studies are somewhat in line with our objectives and methods, they experimented on platforms exhibiting 2-DoF (in roll and pitch) and 3-DoF (in z, roll, and pitch) motions with an 8-DoF quadruped robot. 
In contrast, our research advances to more challenging scenarios, featuring 6-DoF (in x, y, z, roll, pitch, and yaw) platform motions with a 12-DoF quadruped robot for experiments.

\subsection{Privileged Learning and System Identification} 
\label{sec:2-D}
System parameters, also known as privileged information in simulations, are pivotal for learning complex skills using RL~\cite{cheng2023parkour, zhuang2023robot}.
Asymmetric structures facilitate the training of actors by providing critics with the parameters~\cite{nahrendra2023dreamwaq, ma2023learning}. 
Moreover, actors can leverage the parameters by incorporating the system identification process into the RL framework. 
Upon deployments, actors should infer the privileged parameters via online optimization~\cite{yu2018policy, peng2020learning, yu2019sim, yu2020learning} or estimators.
Estimators can predict the parameters from sensor data in three ways: their original form~\cite{Yu2017si, ji2022concurrent}, latent vectors~\cite{kumar2021rma, lee2020learning, fu2023deep, wang2023learning, miki2022learning}, or a hybrid way~\cite{cheng2023parkour}.
Adopting the hybrid representation, we deploy two state estimators: one explicitly estimates the robot and platform states in their original form, and the other implicitly infers the intrinsic properties of the robot within a low-dimensional space. 
These estimations give actors clear situational awareness and domain adaptability, leading to improved balancing performance.
Furthermore, we introduce alignment commands, derived by modulating robot and platform states, for enhanced convergence of the policy.

%% file: Paper/3_method.tex
\begin{figure*}[t]
    \centering 
    \includegraphics[width=1.99\columnwidth]{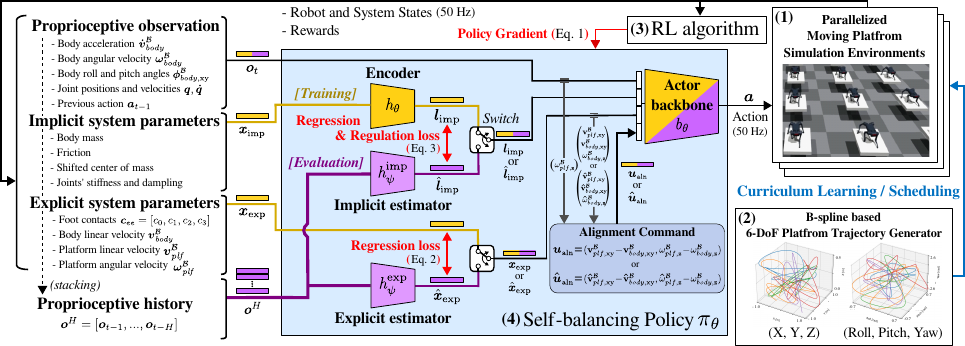}
    \caption{
            \textbf{Overall framework of the Learning-based Active Stabilization method on Moving Platforms (\textit{LAS-MP})}. 
            It consists of four key components: (1) parallelized simulation environments for moving platforms, (2) a platform trajectory generator with a scheduling mechanism for managing task complexity progression, (3) a reinforcement learning (RL) algorithm for policy optimization, and (4) a self-balancing policy with two state estimators.
            To leverage the privileged information in policy learning, we concurrently train the self-balancing policy with the two system state estimators in a single phase by employing the Regularized Online Adaptation (ROA) method~\cite{cheng2023parkour}.
            To clearly differentiate components utilized in training or evaluation phases, we designate yellow for training components and purple for evaluation components. 
            A combination of both colors represents components involved in both phases.
    }
    \vspace{-0.3cm}
    \label{fig:fig2}
\end{figure*}

\section{Learning-based Active Stabilization method on Moving Platforms (\textit{LAS-MP})}
\label{sec:3}
We present the \textit{LAS-MP}, an advanced controller enabling quadruped robots to maintain balance against 6-DoF platform motions without falling off or toppling on the platform.
Below, we discuss the challenges in self-balancing problems on moving platforms along with motivations for the \textit{LAS-MP}, followed by a detailed explanation of each component.

\subsection{Challenges and Motivations}
\label{sec:3-A}
Designing controllers for quadruped robots presents inherent challenges due to their nonlinear, time-varying dynamics encompassing multiple DoF and mechanical constraints~\cite{iqbal2020provably}.
Maintaining balance on moving platforms further demands rapid and suitable responses to external forces on the robot.

Platform motions, represented by continuously changing velocity, expose robots to various forces, including inertia and interaction forces. 
When platforms simultaneously exhibit both translational and rotational motions, the variety of forces increases.
For example, according to Newton's laws of motion, horizontal acceleration induces inertial forces on the robot, sharp turns lead to centrifugal forces, and the platform's rotational motion introduces additional inertial forces around the axis of rotation, like Coriolis, Euler, and centrifugal forces.
In addition, interactions through contacts between the robot and the platform generate normal and frictional forces. 
Thus, upward platform motions can apply vertical reaction forces at contact points, potentially propelling the robot into a flying phase where it loses contact with the platform.
These forces heighten the risk of falling off or toppling on the platform unless using a sophisticated algorithm, equipping with a range of motor skills such as soft landings, contact maintenance, foot-slip prevention, and adaptive postural adjustments for weight redistribution.  

Meanwhile, the lack of information on platform motions causes delayed responses to the forces, requiring more space to regain balance or leading to non-recoverable states.
This is particularly critical with spatial constraints on platforms.

To mitigate these challenges, we introduce the \textit{LAS-MP}. 
In physics simulations, we generate diverse platform motions and apply RL to develop the necessary motor skills. 
We also enhance situational awareness of the policy by inferring robot and platform states in real time using system state estimators, trained via the system identification process, enabling quick and adaptive postural adjustments in the non-inertial frames.
Fig.~\ref{fig:fig2} demonstrates a schematic of the \textit{LAS-MP}.

\begin{figure}[t!]
    \centering 
    \includegraphics[width=\columnwidth]{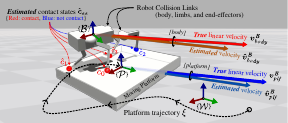}
    \caption{ 
    This figure visually presents the notations used in the descriptions, such as the coordinate frames for the body $\mathcal{B}$, platform $\mathcal{P}$, and world $\mathcal{W}$. 
    It also illustrates estimated robot and platform states, including contact states $\hat{\bm{c}}_{ee}$ and linear velocities of the body $\hat{\bm{v}}^{\mathcal{B}}_{body}$ and platform $\hat{\bm{v}}^{\mathcal{B}}_{plf}$.
    Platform velocities are visualized after conversion to the platform frame for clarity.
    }
    \label{fig:fig3}
    \vspace{-0.45cm}
\end{figure}

\subsection{Variable Notation} 
We first introduce the variable notations to elaborate on the \textit{LAS-MP}. 
As illustrated in Fig.~\ref{fig:fig3}, the locations of the robot and platform are determined by their body $\mathcal{B}$ and platform $\mathcal{P}$ frames, located at their respective center of mass.
Our notation adopts superscripts for coordinate frames and subscripts for specific entities.
For instance, $\bm{p}^{\mathcal{W}}_{\langle \cdot \rangle, [\cdot]} \in \mathbb{R}^3$ denotes the position within the world coordinate frame $\mathcal{W}$, where $\langle \cdot \rangle$ can be $\textit{body}$, $\textit{plf}$, or $\textit{ee}_i$ and $[\cdot]$ indicates specific elements in a variable if used.
For brevity, $\textit{plf}$ and $\textit{ee}_i$ refer to the platform and each leg's end-effector with $i$ ranging from 0 to 3.
$\bm{v}$ and $\bm{\Dot{v}} \in \mathbb{R}^3$ indicate linear velocity and acceleration, while $\bm{\phi}$ and $\bm{\omega} \in \mathbb{R}^3$ represent Euler angles and angular velocity.
$\bm{q}$, $\bm{\Dot{q}}$, and $\bm{\Ddot{q}} \in \mathbb{R}^{12}$ denote joint positions, velocities, and accelerations of the quadruped robot, respectively. 
Feet contact states and forces are represented as $\bm{c}_{ee} = [c_0, c_1, c_2, c_3] \in \mathbb{B}^{4}$ and $\bm{f}_{ee} = [\bm{f}_0, \bm{f}_1, \bm{f}_2, \bm{f}_3] \in \mathbb{R}^{4 \times 3}$.
$\bm{\xi}$ is assigned to the 6-DoF platform trajectory with a total duration of $T$.
Lastly, we use $\hat{\cdot}$ to indicate estimated values in the subsequent descriptions.

\subsection{Problem Formulation of RL}
Our objective of RL is to develop a policy that embodies multiple motor skills required for maintaining balance with minimal deviations on moving platforms. 
We formulate the self-balancing problem as a Partially Observable Markov Decision Process (POMDP) due to the agent's restricted accessibility to privileged system parameters. 
The POMDP is composed of a $7$-tuple $( \mathcal{S}, \mathcal{O}, \mathcal{A}, \mathcal{F}, \mathcal{R}, \mathcal{Q}_0, \gamma )$, where each element is the state space $\mathcal{S}$, the observation space $\mathcal{O} \!\subset\! \mathcal{S}$, the action space $\mathcal{A}$, the transition function $\mathcal{F}: \mathcal{S} \times \mathcal{A} \!\rightarrow\! \mathcal{S}$, the reward function $\mathcal{R}: \mathcal{S} \times \mathcal{A} \!\rightarrow\! \mathbb{R}$, the initial state distribution $\mathcal{Q}_0$, and the discount factor $\gamma \in [0,1)$.
The system parameters, denoted as $\mathcal{X} \subset \mathcal{S}$, comprise elements that are not directly observable through proprioceptive sensors, such as platform velocities and feet friction coefficients. 
For the initial states for each episode $\bm{s}_0 \sim \mathcal{Q}_0$, we position the robot at the center of the platform with a random yaw angle within a [\SI{-180}{\degree}, \SI{180}{\degree}] range.
We then find a skillful self-balancing policy $\pi_\theta$ that maximizes the expected sum of discounted rewards over various platform movements:
\begin{equation}
    J(\pi_\theta) = 
    \mathbb{E}_{ \; \bm{\xi}_{\text{train}} \sim \Xi_{\text{train}}} 
    \left[ 
        \mathbb{E}_{\substack{(\bm{s}, \bm{a}) \sim \rho_{\pi_{\theta}}\\\bm{s}_0 \sim \mathcal{Q}_0}} 
        \left[ 
            \sum_{t=0}^{T} \gamma^t \mathcal{R}(\bm{s}_t, \bm{a}_t)  
        \right] 
    \right],
\label{eq:RLf}
\end{equation}
where $\Xi_{\text{train}}$ is a training set of platform trajectories $\bm{\xi}_{\text{train}}$ and $\rho_{\pi_{\theta}}$ is a state-action visitation probability given a stochastic policy $\pi_\theta$.
In the following description, we detail how RL is implemented in the \textit{LAS-MP}, covering architectures, reward functions, training environments, and technical details.

\subsection{Learning Self-balancing Policy}
\label{sec:3-C}
We employ the Regularized Online Adaptation~\cite{cheng2023parkour} as a privileged learning approach to combine policy optimization and system identification process into a single phase, instead of the two-stage student-teacher approaches~\cite{kumar2021rma, zhuang2023robot}. 
This enables the utilization of the privileged system parameters $\mathcal{X}$ during training in simulation, replacing them with real-time estimates from state estimators upon deployments for evaluations.
Thus, the \textit{LAS-MP} is composed of a self-balancing policy and two system state estimators, as shown in Fig.~\ref{fig:fig2}.

\subsubsection{Self-balancing Policy} 
Our policy $\pi_\theta$ features an actor backbone $b_\theta$ and an encoder $h_{\theta}$, where $\theta$ is network parameters. 
The actor backbone, $b_\theta: \mathcal{O} \times \mathcal{X}_{\text{exp}} \times \mathcal{L}_{\text{imp}} \times se(2) \rightarrow \mathcal{A}$, calculates the action $\bm{a} \in \mathcal{A}$, which is joint displacements $\Delta \bm{q}$, at each timestep. 
The action is combined with nominal joint values, yielding target positions, $\bm{q}_{\text{target}} = \Delta \bm{q} + \bm{q}_\text{nominal}$, for the Proportional Derivative (PD) controllers~\cite{borase2021review}.
For notational clarity, we omit the notation of $t$ here and after. 

The input of the actor backbone $b_\theta$ consists of the observation $\bm{o}\,\in\,\mathcal{O}$, explicit system parameters $\bm{x}_{\text{exp}}\,\in\,\mathcal{X}_{\text{exp}}$, an implicit latent vector $\bm{l}_{\text{imp}}\,\in\,\mathcal{L}_{\text{imp}}$, and an alignment command $\bm{u}_{\text{aln}}\,\in\,se(2)$.  
The observation is a concatenation of proprioceptive sensor data and the previous action, $\bm{o} = [\bm{\Dot{v}}^{\mathcal{B}}_{\text{\textit{body}}}, \bm{\omega}^{\mathcal{B}}_{\text{\textit{body}}}, \bm{\phi}^{\mathcal{B}}_{\text{\textit{body},xy}}, \bm{q}, \bm{\Dot{q}}, \bm{a}_{t-1}]$. 
For the privileged system parameters $\mathcal{X}$, we categorize them into two subgroups: explicit and implicit ones, $\mathcal{X}_{\text{exp}}$ and $\mathcal{X}_{\text{imp}}$. 
The explicit system parameters $\bm{x}_{\text{exp}}\,\in\,\mathcal{X}_{\text{exp}}$ include foot contacts and body linear velocity, and platform's linear and angular velocities, $\bm{x}_{\text{exp}} = [\bm{c}_{ee}, \bm{v}^{\mathcal{B}}_{\textit{body}}, \bm{v}^{\mathcal{B}}_{\textit{plf}}, \bm{\omega}^{\mathcal{B}}_{\textit{plf}}]$. 
This input allows the policy to recognize the motions of the reference frame, enabling adaptive postural adjustments in response to platform motions.
The implicit system parameters $\bm{x}_{\text{imp}}\,\in\,\mathcal{X}_{\text{imp}}$, representing intrinsic properties of the robot, consist of four coefficients: $\bm{x}_{\text{imp}} = [$body mass, shifted center of mass (CoM), contact friction, stiffness and damping of joints$]$, encoded into the latent vector $\bm{l}_{\text{imp}}$ by the encoder $h_{\theta}: \mathcal{X}_{\text{imp}} \rightarrow \mathcal{L}_{\text{imp}}$. 
This vector helps the policy adapt to different intrinsic properties in deployments~\cite{kumar2021rma}.
The last input, the alignment command $\bm{u}_{\text{aln}} \!=\! [\bm{v}^{\mathcal{B}}_{\text{\textit{plf},xy}} \! - \! \bm{v}^{\mathcal{B}}_{\text{\textit{body},xy}}, \omega^{\mathcal{B}}_{\text{\textit{plf},z}} \! - \! \omega^{\mathcal{B}}_{\text{\textit{body},z}}]$, is engineered to indicate the robot's relative velocity to the platform. 
This functions as a body velocity command, guiding the policy to correct misalignment between the robot and the platform.

\subsubsection{System State Estimators} 
In deployments, since privileged information is inaccessible, we substitute the input elements of the actor backbone related to the privileged information with real-time estimates derived from the history of observations, $\bm{o}^{H} = [\bm{o}_{t-1}, ..., \bm{o}_{t-H}] \in \mathcal{O}^{H}$.
To realize that, we employ explicit and implicit state estimators. 
The explicit estimator, $h^{\text{exp}}_{\psi}: \mathcal{O}^{H} \!\rightarrow\! \mathcal{X}_{\text{exp}}$, estimates the explicit system parameter, $\hat{\bm{x}}_{\text{exp}} = [\hat{\bm{c}}_{ee}, \hat{\bm{v}}^{\mathcal{B}}_{\textit{body}}, \hat{\bm{v}}^{\mathcal{B}}_{\textit{plf}}, \hat{\bm{\omega}}^{\mathcal{B}}_{\textit{plf}}]$, as depicted in Fig.~\ref{fig:fig3}.
We then adopt the estimated explicit parameters to generate the inferred alignment command, $\hat{\bm{u}}_{\text{aln}} = [\hat{\bm{v}}^{\mathcal{B}}_{\text{\textit{plf},xy}} - \hat{\bm{v}}^{\mathcal{B}}_{\text{\textit{body},xy}}, \hat{\omega}^{\mathcal{B}}_{\text{\textit{plf},z}} - \omega^{\mathcal{B}}_{\text{\textit{body},z}}]$.
For the z-directional body angular velocity $\omega^{\mathcal{B}}_{\text{\textit{body},z}}$, we use the measured value by accessible inertial measurement unit (IMU) sensors.
The implicit state estimator, $h^{\text{imp}}_{\psi}: \mathcal{O}^{H} \!\rightarrow\! \mathcal{L}_{\text{imp}}$, estimates the implicit latent vector $\hat{\bm{l}}_{\text{imp}}$ which represents the inferred intrinsic properties of the robot. 
The network parameters of the state estimators, denoted as $\psi$, are trained concurrently with the RL objective function (Eq.~\ref{eq:RLf}) using Mean Squared Error (MSE) losses: 
\begin{align}
L^{\text{exp}}_{\text{\scriptsize MSE}} &= || \hat{\bm{x}}_{\text{exp}} - \bm{x}_{\text{exp}} ||_2^2, \\
L^{\text{imp}}_{\text{\scriptsize MSE}} &= || \hat{\bm{l}}_{\text{imp}} - sg[\bm{l}_{\text{imp}}] ||_2^2 + \lambda || sg[\hat{\bm{l}}_{\text{imp}}] - \bm{l}_{\text{imp}} ||_2^2,
\label{eq:Rgf}
\end{align}
where $sg[\cdot]$ is a stop gradient operator and $\lambda$ is a Lagrangian multiplier. 
The first term in each loss function is for regression of each estimator, while the second term of Eq.~\ref{eq:Rgf} aims for regularization of the encoder $h_\theta$ within the policy $\pi_\theta$.

\begin{table}[t]
\centering
\begin{threeparttable}
\caption{Reward Function of RL: $\mathcal{R} = \mathcal{R}^{\text{task}} + \mathcal{R}^{\text{reg}}$ \\ 
(Please refer to the description in Sec.~\ref{sec:3-C-3} for more details.)}
\label{table:table0}
\renewcommand{\arraystretch}{1.3}
\begin{tabular}{c|c}
\hline
\textbf{Reward} & \textbf{Function Expression} \\
\hline
\multicolumn{2}{c}{Task Reward: $\mathcal{R}^{\text{task}} = \sum_{k=0}^{4} r^{\text{task}}_{k}$} \\
\hline
$r^{\text{task}}_{0}$ & $-k_0 \, \mathds{1}_{\text{collision}}(\bm{q}, \bm{p}^{\mathcal{P}}_{\textit{body}}, \bm{\phi}^{\mathcal{P}}_{\textit{body}})$ \\
$r^{\text{task}}_{1}$ & $k_1 \, \exp(-\| \bm{p}^{\mathcal{P}}_{\text{body,xy}} \|_2 / k_2)$ \\
$r^{\text{task}}_{2}$ & $k_3 \, \exp(-\| \bm{u}_{\text{aln}} \|_2 / k_4)$ \\
$r^{\text{task}}_{3}$ & $k_5 \, \exp(-\| \bm{\phi}^{\mathcal{W}}_{\text{body,xy}} \|_2 / k_6)$ \\
$r^{\text{task}}_{4}$ & $k_7 \, \exp(-\| p^{\mathcal{P}}_{\text{body,z}} - h^{\textit{body}}_{\text{des}} \|_2 / k_8)$ \\
\hline
\multicolumn{2}{c}{Regularization Reward: $\mathcal{R}^{\text{reg}} = \sum_{k=0}^{6} r^{\text{reg}}_{k}$} \\
\hline
$r^{\text{reg}}_{0}$ & $- (k_{9} \, \| \bm{\tau} - \bm{\tau}_{t-1} \|_2 + k_{10} \, \| \bm{a} - \bm{a}_{t-1} \|_2) $\\
$r^{\text{reg}}_{1}$ & $- (k_{11} \, \| \bm{\tau} \|_2 + k_{12} \, \| \bm{\dot{q}} \|_2 + k_{13} \, \| \bm{\ddot{q}} \|_2) $ \\
$r^{\text{reg}}_{2}$ & $- k_{14} \, \sum_{j=0}^{11} \max( \tau_j \dot{q}_j, 0.0) $ \\
$r^{\text{reg}}_{3}$ & $- k_{15} \sum_{i=0}^{3} \max(\| \bm{f}_{i} \|_2 - f_{\text{tol}}, 0.0) $ \\
$r^{\text{reg}}_{4}$ & $- k_{16} \sum_{i=0}^{3} (t^{\text{swing}}_{i} - t^{\text{swing}}_{\text{des}})(c_i (1-c_{i, t-1}))$ \\
$r^{\text{reg}}_{5}$ & $- k_{17} \sum_{i=0}^{3} (t^{\text{contact}}_{i} - t^{\text{contact}}_{\text{des}})((1 - c_i) c_{i, t-1}) $ \\
$r^{\text{reg}}_{6}$ & $- k_{18} \sum_{i=0}^{3} (\| \bm{v}^{\mathcal{P}}_{\text{ee}_i\text{,xy}} - ( \omega^{\mathcal{P}}_{\text{plf,z}} \bm{p}^{\mathcal{P}}_{\text{ee}_i\text{,xy}} + \bm{v}^{\mathcal{P}}_{\text{plf,xy}}) \|_2)$ \\
\hline
\end{tabular}
\begin{tablenotes}[flushleft] 
    \scriptsize
    \item \textbullet\; $\mathds{1}$: an indicator function, $k_{0, \dots, 18}$: non-negative coefficients.
    \item \textbullet\; $h^{\text{body}}_{\text{des}}$: a desired body height, $f_{\text{tol}}$: a maximum tolerated contact force.
    \item \textbullet\; $t^{\text{swing}}_{\text{des}}$: a desired contact duration, $t^{\text{contact}}_{\text{des}}$: a desired swing duration.
\end{tablenotes}
\end{threeparttable}
\vspace{-0.5cm}
\end{table}

\subsubsection{Reward Function} 
\label{sec:3-C-3}
To develop the essential motor skills using RL for self-balancing tasks on moving platforms, we design the reward function combining task and regularization rewards, represented as $\mathcal{R} = \mathcal{R}^{\text{task}} + \mathcal{R}^{\text{reg}}$.
TABLE~\ref{table:table0} shows the detailed composition of each reward function.

We form the task reward $\mathcal{R}^{\text{task}}=\sum_{k=0}^{4} r^{\text{task}}_{k}$ to encompass the necessary aspects of solving the task. 
The $r^{\text{task}}_{0}$ penalizes collisions to ensure safety, the $r^{\text{task}}_{1}$ minimizes positional deviations to prevent falls, the $r^{\text{task}}_{2}$ aligns the robot's motion with the platform, and the $r^{\text{task}}_{3}$ reduces body tilt for postural stabilization.
The $r^{\text{task}}_{4}$ ensures that the robot maintains its desired body height to preserve its functionality, such as locomotion and manipulation, rather than just lying down.

Adhering to only the task reward might lead the agent into local minima, resulting in unnatural motion and excessive power usage~\cite{Suyoung2023deformter, ji2022concurrent}. 
To mitigate this, we adopt the regularization reward $\mathcal{R}^{\text{reg}}= \sum_{k=0}^{6} r^{\text{reg}}_{k}$.  
The $r^{\text{reg}}_{0}$ and $r^{\text{reg}}_{1}$ reduce jerky robot motion, and the $r^{\text{reg}}_{2}$ lessens power consumption. 
The $r^{\text{reg}}_{3}$ aims to regulate the magnitude of contact forces, and the $r^{\text{reg}}_{4}$ and $r^{\text{reg}}_{5}$ discourage frequent contact and lifting of the feet.
Foot replacements should be performed only when required to get a re-balance to prevent deviations caused by the platform's independent movements.
Lastly, we propose the $r^{\text{reg}}_{6}$ to penalize the foot slip on the moving platform by taking the underlying platform's motions into account.

\subsubsection{Training Environments}
\label{sec:3-C-4}
We formed parallel simulation environments for efficient data collection during the training phase; each environment consisted of a quadruped robot and a box-shaped moving platform as agents.
Each episode starts by positioning the robot on the platform according to the initial state distribution $\mathcal{Q}_0$. 
To improve generalization for varying intrinsic properties of the robot, the intrinsic parameters $\bm{x}_{\text{imp}}$ are randomly sampled as described in TABLE~\ref{table:table1}.

To realize platform motions, we parameterized the shape of the platform with [width, length, height] (unit: \SI{}{\meter}), and the PD controllers are used to track the 6-DoF training platform trajectories $\bm{\xi}_{\text{train}} \in \Xi_{\text{train}}$.
We diversified platform motions by equipping each environment with trajectory generators and randomizing controller parameters for every episode.
These generators craft a training set of platform trajectories $\Xi_{\text{train}}$ utilizing B-spline interpolation~\cite{knott1999interpolating} with randomly chosen waypoints in the ranges: x and y: $[-1.0, 1.0]$ \SI{}{\meter}, z: $[0.0, 5.0]$ \SI{}{\meter}, roll and pitch: $[-0.7, 0.7]$ \SI{}{\radian}, and yaw: $[-2.6, 2.6]$ \SI{}{\radian}, respectively. 
Each trajectory spans for $T=10$ \SI{}{\second}.

\begin{table}[t]
\centering
\caption{Training and Testing Parameter Ranges}
\renewcommand{\arraystretch}{1.2}
\begin{tabular}{c|c|c}
\hline
\textbf{Parameters} & \textbf{Training Range} & \textbf{Testing Range} \\
\hline
\multicolumn{3}{c}{Implicit System Parameters $\bm{x}_{\text{imp}}$}  \\
\hline
Body mass (\SI{}{\kilogram})                  & $U(4.0, 5.0)$         & $U(3.5, 5.5)$ \\
Shifted CoM (\SI{}{\meter})     & $U(-0.2, 0.2)^{3}$    & $U(-0.25, 0.25)^{3}$ \\
Contact friction                        & $U(0.8, 1.2)$         & $U(0.7, 1.3)$ \\
Stiffness ($K_p$)               & $U(36, 44)^{12}$      & $U(32, 48)^{12}$ \\
Damping ($K_d$)                 & $U(0.8, 1.2)^{12}$    & $U(0.6, 1.4)^{12}$ \\
\hline
\multicolumn{3}{c}{Platform Trajectory $\bm{\xi}$}  \\
\hline
Number of waypoints  & $[5, 6, ..., 15]$           & $[4, 5, ..., 16]$ \\
Stiffness ($K_p$)   & $U(1.0, 1.5)^6$           & $U(0.5, 2.0)^6$ \\
Damping ($K_d$)     & $U(0.02, 0.03)^6$         & $U(0.01, 0.04)^6$ \\
\hline
\end{tabular}
\label{table:table1}
\begin{tablenotes}
\setlength{\itemindent}{0.1cm}
\item[] \textbullet\; $U(\cdot, \cdot)$: an uniform distribution
\end{tablenotes}
\vspace{-0.6cm}
\end{table}

\begin{figure}[t]
    \centering 
    \includegraphics[width=\columnwidth]{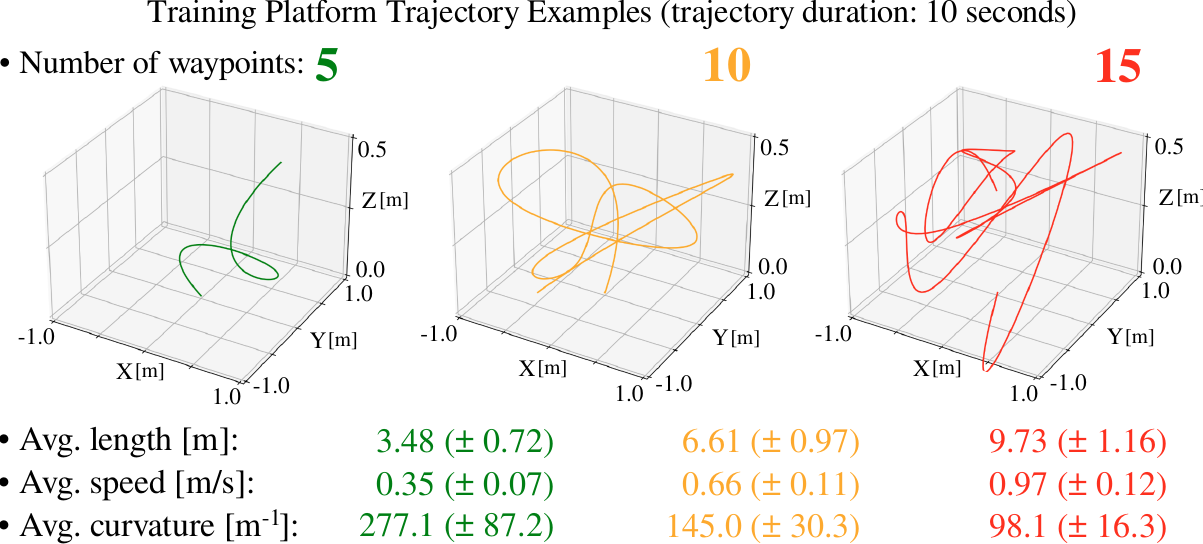}
	\caption{
        Examples of training platform trajectories $\Xi_{\text{train}}$ in translational space and their statistical analysis.
        A similar trend is observed in rotational space.
	}
        \label{fig:fig4}
        \vspace{-0.5cm}
\end{figure}

Figure~\ref{fig:fig4} shows translational platform trajectory examples and provides statistical analysis for the $\Xi_{\text{train}}$. 
Adding more waypoints while maintaining a constant duration of \SI{10}{\second} led to increased path lengths and higher speeds on average. 
Rapid direction changes at high speeds posed significant balancing challenges for the robot, especially at the early stages of training, hindering the collection of meaningful data for policy training. 
To successfully train the policy, we adopted curriculum learning methods~\cite{margolis2022rapid, kumar2021rma, cheng2023parkour}, setting task complexity based on the number of waypoints as shown in TABLE~\ref{table:table1}. 
A scheduler then incrementally escalates the task level after achieving an 80\% success rate at each level.

\subsubsection{Training Details}
\label{sec:3-C-5}
We adopted the Proximal Policy Optimization~\cite{schulman2017proximal} with ROA regularization~\cite{cheng2023parkour}, aiming to maximize the RL objective function (Eq.~\ref{eq:RLf}) and minimize the regression and regularization losses (Eq.~2, \ref{eq:Rgf}). 
The actor backbone $b_\theta$ is constructed using a four-layer Multi-Layer Perceptron (MLP) network, while the encoder $h_{\theta}$ is implemented as a two-layer MLP. 
Concurrently, both the explicit and implicit state estimators, $h^{\text{exp}}_{\psi}$ and $h^{\text{imp}}_{\psi}$, are designed with a two-layer 1D Convolutional Neural Network (CNN) in between two one-layer MLPs.
The stochastic policy $\pi_\theta$ is conceptualized as a diagonal Gaussian distribution, wherein the mean value is generated by the actor backbone $b_\theta$ with a parameterized standard deviation $\bm{\theta}_{std} \in \mathbb{R}^{12}$.
Then, actions are sampled following the distribution, $\bm{a} \sim \mathcal{N}(b_\theta, \bm{\theta}_{std})$.

We empirically found the best performing reward coefficients $k_{0, ...,18}$ as [$10.0$, $3.0$, $1.2$, $2.0$, $0.3$, $1.0$, $0.2$, $4.0$, $0.1$, $10^{-7}$, $10^{-4}$, $10^{-4}$, $10^{-7}$, $10^{-6}$, $10^{-5}$, $0.01$, $2.0$, $3.0$, $0.05$]. 
We also set the parameters $H$, $\lambda$, $h^{\text{body}}_{\text{des}}$, $f_{\text{tol}}$, $t^{\text{swing}}_{\text{des}}$, and $t^{\text{contact}}_{\text{des}}$ as 20, 0.2, \SI{0.37}{\m}, \SI{50}{\N}, \SI{0.1}{\s}, and \SI{0.5}{\s}, respectively.

We utilized the Isaac Gym~\cite{makoviychuk2021isaac} to operate $8,192$ simulation environments in parallel.
As a quadruped robot, we adopted a Unitree A1~\cite{wang2024unitree}, which weights $11.74$ \SI{}{\kilogram} and has dimensions of $48$ \SI{}{\cm} in length, $32$ \SI{}{\cm} in width, and $37$ \SI{}{\cm} in height at the nominal standing configuration $\bm{q}_{\text{nominal}}$.
Additionally, we configured the shape of the platforms as [$2.0, 2.0, 0.2$] \SI{}{\m}. 
The total training consisted of $4,000$ iterations, with each iteration processing data equivalent to $0.24$ \SI{}{\s}.
This equates to $96$ trajectories $\bm{\xi}_{\text{train}}$ per environment, which is about $0.8$ million trajectories $\Xi_{\text{train}}$ when summed across all environments.
The overall training required about $2$ hours and $30$ minutes, executed on a standard desktop configuration with an Intel i9-9900K CPU and an RTX 4090 GPU.

%% file: Paper/4_experiment.tex
\section{Experimental Results}
\label{sec:4}
The major advantage of the \textit{LAS-MP} lies in its balancing proficiency across various platform motions. 
In experiments, to corroborate the proficiency, we compare the \textit{LAS-MP} with three other baseline methods.
We also perform ablation studies to validate the impacts of platform state estimation using explicit estimators (\textit{EE}) and of alignment commands (\textit{AC}).
Lastly, we assess the system state estimators to demonstrate their feasibility in inferring system states using a history of proprioceptive sensor data.

\begin{figure*}[t]
    \centering 
    \includegraphics[width=2\columnwidth]{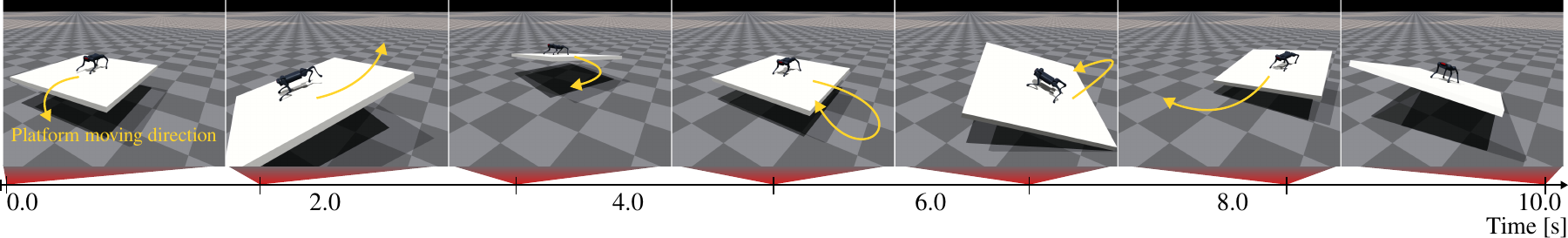}
    \vspace{-0.1cm} 
    \caption{ 
    Snapshots undergoing an evaluation process of the \textit{LAS-MP}, captured with a fixed camera in the world coordinate frame $\mathcal{W}$.
    }
    \label{fig:fig5}
    \vspace{-0.15cm}
\end{figure*}

\begin{table*}[ht]
\centering
\caption{Balancing Performance on Evaluation Benchmark Set $\Xi_{\text{eval}}$ consisting of $10,000$ Evaluation Platform Trajectories $\xi_{\text{eval}}$}
\renewcommand{\arraystretch}{1.3}
\begin{tabular}{l|cc|cc|c}
\hline 
\multicolumn{1}{c|}{\multirow{2}{*}{\textbf{Method}}} & \multicolumn{2}{c|}{\textbf{Constraint violation rate}} & \multicolumn{2}{c|}{\textbf{Deviations from initial pose}} & \multirow{2}{*}{\textbf{Power} (\SI{}{\watt}) $\downarrow$} \\ \cline{2-5}
& Collision (\%*) $\downarrow$ & Height (\%*) $\downarrow$ & Position (\SI{}{\m}) $\downarrow$ & Rotation (\SI{}{\radian}) $\downarrow$ & \\
\hline 
\textit{LAS-MP (ours)}                       & \textbf{0.029 (±0.03)} & \textbf{0.009 (±0.01)} & \textbf{0.366 (±0.01)} & \textbf{0.071 (±0.01)} & 2.904 (±0.45) \\ 
\hline
\textit{Stand Still}                         & 0.612 (±0.29) & 0.142 (±0.07) & 0.394 (±0.04) & 0.182 (±0.07) & \textbf{1.031 (±0.17)} \\ 
\hline
\textit{Rough-Static Policy (R-S Policy)}~\cite{cheng2023parkour}   & 0.922 (±0.12) & 0.232 (±0.07) & 0.478 (±0.06)  & 0.241 (±0.05) & 8.121 (±2.86) \\             
\textit{R-S Policy w/ Oracle Command}             & 0.891 (±0.18) & 0.198 (±0.01) & 0.452 (±0.05) & 0.201 (±0.04) & 7.817 (±2.95) \\
\hline
\multicolumn{1}{c}{ \multirow{2}{*}{\textbf{Ablation Study}} } & \multicolumn{5}{r}{ \multirow{2}{*}{\textbullet \, \%*: a normalized percentage in the range of [0.0, 1.0]} } \\
\multicolumn{6}{c}{ } \\
\hline
\textit{w/o Alignment Command (AC)}              & 0.287 (±0.27) & 0.327 (±0.12) & 0.367 (±0.05) & 0.195 (±0.08) & 4.658 (±1.55) \\
\textit{w/o EE\{PLV, PAV\} \& w/o AC}            & 0.634 (±0.37) & 0.454 (±0.29) & 0.376 (±0.04) & 0.122 (±0.06) & 3.388 (±0.83) \\
\textit{w/o EE \& w/o AC}                        & 0.359 (±0.13) & 0.534 (±0.16) & 0.496 (±0.03) & 0.159 (±0.04) & 6.614 (±2.63) \\
\textit{w/o EE \& w/o AC \& w/ History Obs.}     & 0.314 (±0.19) & 0.525 (±0.29) & 0.456 (±0.07) & 0.177 (±0.09) & 6.139 (±4.36) \\
\hline
\addlinespace
\multicolumn{6}{r}{\textbullet \, \textit{EE}: Explicit Estimator, \textit{PLV}: Platform Linear Velocity, \textit{PAV}: Platform Angular Velocity, \textit{Obs.}: Observations}  \\
\end{tabular}
\label{table:table2}
\vspace{-0.55cm}
\end{table*}

\subsection{Evaluation of Balancing Performance}
\subsubsection{Experimental Setup}
We aim to evaluate the balancing and domain adaptation capabilities across diverse platform motions and robot intrinsic properties, using a broader range of parameters than those used in training, as shown in TABLE~\ref{table:table1}.
To this end, we created an evaluation benchmark set $\Xi_{\text{eval}}$ consisting of $10,000$ evaluation platform trajectories $\xi_{\text{eval}}$, following the same procedure in Sec.~\ref{sec:3-C-4} with the random number of waypoints within the testing range.
This benchmark statistically shows a length of $7.12$ \SI{}{\m}, a speed of $0.69$ \SI{}{\meter}/\SI{}{\s}, and a curvature of $132.6$ \SI{}{\per\m} on average in translational space.
Fig.~\ref{fig:fig5} displays an evaluation episode of the \textit{LAS-MP} alongside 6-DoF platform movements examples.
We also prepared 1,024 quadruped robots having different intrinsic properties for each robot.

For evaluation, we organized four metrics: (1) the collision rate on the body and limbs excepting the end-effectors (i.e., feet), (2) the rate of height constraint violations beyond the $\pm \, 0.1$ \SI{}{\meter} range from the desired body height $h^{\text{body}}_{\text{des}}$, (3) the deviations from initial poses, and (4) the power consumption.
Note that we consider falling off the platform as a collision, and height constraint violations and deviations are measured up until the first collision happens.

To demonstrate the effectiveness of our active stabilization method, we prepared the following baselines for comparison:

\begin{itemize}[leftmargin=0.55cm]
    \item \textit{LAS-MP (ours)}: It is our proposed method in Sec.~\ref{sec:3}.
    \item \textit{Stand Still}:
    It keeps the robot in a standing posture by setting the nominal configuration $\bm{q}_{\text{nominal}}$ as the target of the PD controller.
    \item \textit{Rough-Static Policy (R-S Policy)}~\cite{cheng2023parkour}:
    It is a robust locomotion policy trained on static and rough terrains with push perturbations on the body. 
    While structurally similar to our method, having explicit and implicit estimators, it does not consider the platform states. 
    Also, the policy is not trained on the moving platform environments.
    In the experiments, x, y, and yaw velocity commands are set to zero, reflecting the assumption that no prior knowledge is given about the platform's movements.
    \item \textit{R-S Policy w/ Oracle Command}:
    It is designed to augment the \textit{R-S Policy} with Oracle information, such as positional information of the robot and platform in the world coordinate frame $\mathcal{W}$.
    The policy takes the velocity commands, aimed toward the center of the platform, as input. 
    It helps the robot align with the platform movements and compensate for deviations via locomotion.
\end{itemize}
We prepared the \textit{Stand Still} to verify if the intrinsic structural stability of four-legged robots is adequate for maintaining balance on moving platforms without active adjustments. 
In addition, we explored the \textit{R-S Policy} to determine if policies trained exclusively in static environments can sustain balance on dynamically moving platforms.

\subsubsection{Experimental Result}
TABLE~\ref{table:table2} shows the balancing performance of each method on the $\Xi_{\text{eval}}$ benchmark set. 
The \textit{LAS-MP} stands out with superior balancing performance, showing minimal constraint violations and deviations.
On the other hand, the \textit{Stand Still} shows the least power usage with a reasonable height constraint violation rate, since it remains the standing posture. 
However, its high collision rate indicates its inadequacy in maintaining balance on the moving platforms, showing the necessity for advanced balancing strategies.
Furthermore, the \textit{Rough-Static Policy} exhibits the highest collision rate, and even with the Oracle information, only limited improvement is observed.
This result indicates the existence of distinct requirements for getting the balance on moving platforms compared to static environments.

\begin{table*}[ht]
\centering
\caption{Prediction Accuracy of Explicit ($h^{\text{exp}}_{\psi}$) and Implicit ($h^{\text{imp}}_{\psi}$) State Estimators within the \textit{LAS-MP} on the $\Xi_{\text{eval}}$ Benchmark Set}
\vspace{-0.1cm}
\renewcommand{\arraystretch}{1.5}
\resizebox{2\columnwidth}{!}{
    \begin{tabular}{cccc|ccc|ccc|ccc|c}
    \hline
    \multicolumn{13}{c|}{$h^{\text{exp}}_{\psi}$: $|| \hat{\bm{x}}_{\text{exp}, i} - \bm{x}_{\text{exp}, i}||_1 (i=[0, 1, ..., 12])$ } & $h^{\text{imp}}_{\psi}$:  \\
    \cline{1-13}
    \multicolumn{4}{c|}{Foot Contacts $\in \mathbb{R}^{4}$} & \multicolumn{3}{c|}{Body Linear Velocity $\in \mathbb{R}^{3}$} & \multicolumn{3}{c|}{Platform Linear Velocity $\in \mathbb{R}^{3}$} & \multicolumn{3}{c|}{Platform Angular Velocity $\in \mathbb{R}^{3}$} &  $|| \hat{\bm{l}}_{\text{imp}} - \bm{l}_{\text{imp}} ||_2$\\
    \hline
    0.074 & 0.077 & 0.081 & 0.090 & 0.035 & 0.047 & 0.038 & 0.040 & 0.051 & 0.035 & 0.027 & 0.028 & 0.016 & 0.0059 \\
    (±0.019) & (±0.012) & (±0.008) & (±0.016) & (±0.005) & (±0.005) & (±0.002) & (±0.005) & (±0.005) & (±0.002) & (±0.003) & (±0.003) & (±0.002) & (±0.0018)\\
    \hline
    \end{tabular}
    }
\vspace{-0.3cm}
\label{table:table3}
\end{table*}

 \begin{figure}[t!]
    \centering 
    \vspace{-0.1cm}
    
    \includegraphics[width=\columnwidth]{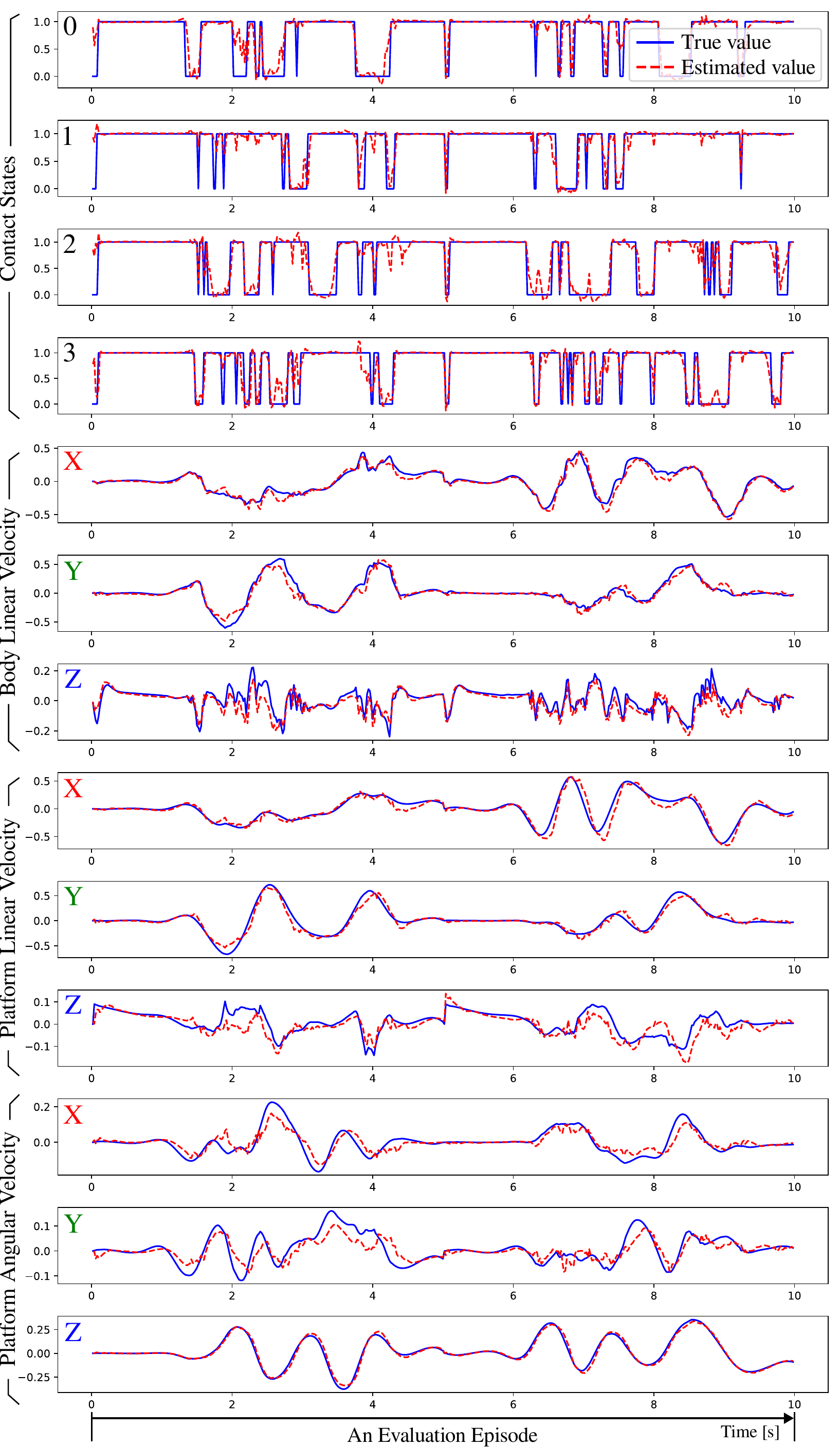}
	\caption{
        This figure presents the predicted $\hat{\bm{x}}_{\text{exp}}$ (depicted as red dotted lines) and ground-truth $\bm{x}_{\text{exp}}$ (shown as blue solid lines) values for each explicit parameter over the course of an evaluation episode spanning \SI{10}{\sec}.
	}
        \label{fig:fig6}
        \vspace{-0.8cm}
\end{figure}

\subsection{Ablation Study}
We carried out ablation studies to assess the efficacy of platform state estimation via the explicit estimator (\textit{EE}) and of the alignment command (\textit{AC}) within the \textit{LAS-MP}. 
Note that in this experiment, we kept the environment setting and learning process consistent with the approach outlined in Sec.~\ref{sec:3}, varying only in the model architecture.
TABLE~\ref{table:table2} shows the balancing performance of each architecture. 
In the \textit{w/o Alignment Command (AC)} case, it shows the policy converged to a local minimum without the \textit{AC}, indicating a performance drop compared to the \textit{LAS-MP}.
It represents the \textit{AC} helps the policy converge to a better local minimum by explicitly extracting task-relevant features. 
As mentioned in~\cite{lesort2018state}, feature engineering is crucial for policy learning.
In the \textit{w/o EE\{PLV, PAV\} \& w/o AC} study, we adjusted the explicit estimator (\textit{EE}) only to infer the robot-related states, excluding the platform linear velocity (\textit{PLV}) and the platform angular velocity (\textit{PAV}).
This is the same with the \textit{R-S Policy}~\cite{cheng2023parkour} but trained on moving platform environments. 
However, it shows higher constraint violation rates.
The robot's linear velocity alone, influenced by both robot and platform motions, failed to provide a clear situation awareness in non-inertia frames, introducing noise into the decision-making process.
In the \textit{w/o EE \& w/o AC} case, we completely omitted the usage of the explicit estimator. 
Interestingly, it shows improved performance in the collision rate with high power consumption. 
The converged policy developed a jumping strategy to effectively ignore the diverse disturbances from platforms moving in any direction.
However, this naturally led to the increased height constraint violation rate, power usage, and positional deviation. 
Lastly, in the case of \textit{w/o EE \& w/o AC \& w/ History Obs.}, aiming for a fair comparison, we utilized historical proprioceptive sensor data as input for the policy. 
This resulted in a slight improvement in overall balancing performance. 
Nonetheless, the policy still converged to the same jumping strategy.

\begin{figure}[t!]
    \centering 
    \includegraphics[width=\columnwidth]{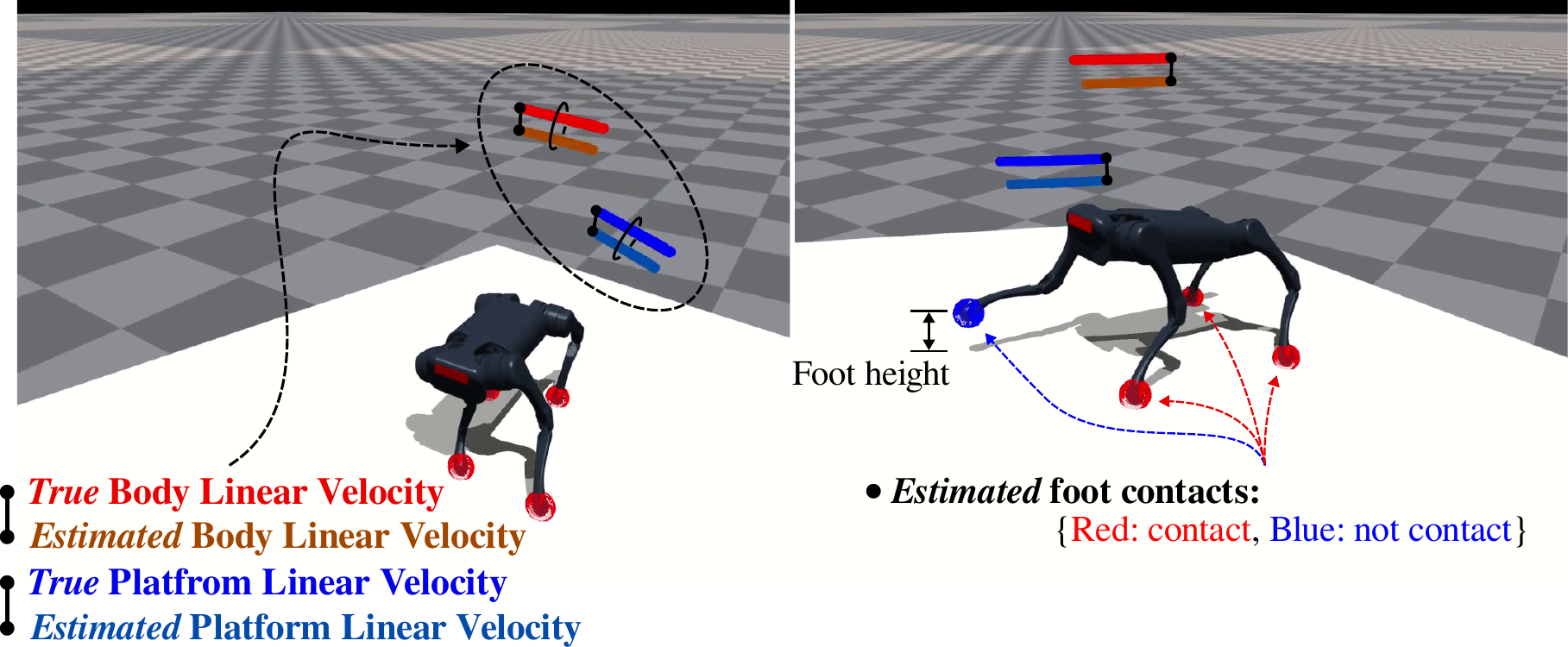}
	\caption{
Snapshots, captured during the evaluation of the \textit{LAS-MP}, show the estimated states from the explicit estimator.
To clearly display each velocity, we draw them by moving a certain distance along the z-axis from the center of both the robot and the platform in the world coordinate frame $\mathcal{W}$.
	}
        \label{fig:fig7}
        \vspace{-0.7cm}
\end{figure}

\subsection{Analysis of State Estimators}
We further examined the accuracy of estimation results of each estimator within the \textit{LAS-MP}.
TABLE~\ref{table:table3} shows the prediction accuracy of the explicit ($h^{\text{exp}}_{\psi}$) and implicit ($h^{\text{imp}}_{\psi}$) estimators.
Accuracy is measured by $|| \hat{\bm{x}}_{\text{exp}, i} - \bm{x}_{\text{exp}, i}||_1$ for each explicit parameter $i = [0, 1, ..., 12]$ and $|| \hat{\bm{l}}_{\text{imp}} - \bm{l}_{\text{imp}} ||_2$ for implicit parameters.
The explicit estimator shows quite low prediction errors for each parameter, indicating it can regress the robot and platform states using historical proprioceptive data. 
The accuracy of the implicit estimator also exhibits its ability to identify the robot's intrinsic properties during the balancing act. 
Fig.~\ref{fig:fig6} and~\ref{fig:fig7} display estimation results of the explicit estimator $h^{\text{exp}}_{\psi}$ on some representative situations. 

\vspace{0.1cm}
For a more intuitive understanding of the experimental results, please refer to the supplementary video.

%% file: Paper/6_conclusion.tex
\section{Conclusion}
\label{sec:5}
We have presented the Learning-based Active Stabilization Method on Moving Platforms (\textit{LAS-MP}).
Through a series of experiments, we have demonstrated that the \textit{LAS-MP} significantly enhances balancing performance on moving platforms exhibiting 6-DoF movements. 
In addition, experiments have shown that disturbances generated by dynamically moving platforms present a potential safety risk to the baseline systems.
In future works, we will extend this work to real-world experiments to verify the feasibility. 
In addition, although the \textit{LAS-MP} aligns the robot motions with those of the platform by generating instantaneous alignment commands, it lacks a mechanism to correct already established steady-state deviations. 
Therefore, we plan to localize the center of the platform and reduce deviations via locomotion by utilizing exteroceptive sensor data.

\section*{Acknowledgments}
\footnotesize
This work was supported by grants from the Korea government (MSIT), including the Institute of Information \& communications Technology Planning \& Evaluation (IITP) grant (No. RS-2023-00237965) and the National Research Foundation of Korea (NRF) grant (No. RS-2023-00208506).

%% file: ref.bib
@article{delmerico2019current,
  title={The current state and future outlook of rescue robotics},
  author={Delmerico, J. and Mintchev, S. and Giusti, A. and Gromov, B. and Melo, K. and Horvat, T. and Cadena, C. and Hutter, M. and others},
  journal={Journal of Field Robotics},
  volume={36},
  number={7},
  pages={1171--1191},
  year={2019},
  publisher={Wiley Online Library}
}

@article{bellicoso2018advances,
  title={Advances in real-world applications for legged robots},
  author={Bellicoso, D. and Bjelonic, M. and Wellhausen, L. and Holtmann, K. and Gunther, F. and Tranzatto, M. and Fankhauser, P. and Hutter, M.},
  journal={Journal of Field Robotics},
  volume={35},
  number={8},
  pages={1311--1326},
  year={2018},
  publisher={Wiley Online Library}
}

@book{knott1999interpolating,
  title={Interpolating cubic splines},
  author={Knott, G. D},
  volume={18},
  year={1999},
  publisher={Springer Science \& Business Media}
}

@inproceedings{fankhauser2018robust,
  title={Robust rough-terrain locomotion with a quadrupedal robot},
  author={Fankhauser, P. and Bjelonic, M. and Bellicoso, C. D. and Miki, T. and Hutter, M.},
  booktitle={Proceedings of the IEEE International Conference on Robotics and Automation (ICRA)},
  pages={5761--5768},
  year={2018},
  organization={IEEE}
}

@article{mastalli2022agile,
  title={Agile maneuvers in legged robots: a predictive control approach},
  author={Mastalli, C. and Merkt, W. and Xin, G. and Shim, J. and Mistry, M. and Havoutis, I. and Vijayakumar, S.},
  journal={arXiv preprint arXiv:2203.07554},
  year={2022}
}

@article{lu2023whole,
  title={Whole-body motion planning and control of a quadruped robot for challenging terrain},
  author={Lu, G. and Chen, T. and Rong, X. and Zhang, G. and Bi, J. and Cao, J. and Jiang, H. and Li, Y.},
  journal={Journal of Field Robotics},
  pages={1657--1677},
  year={2023}
}

@ARTICLE{9779429,
  title={RLOC: Terrain-Aware Legged Locomotion Using Reinforcement Learning and Optimal Control}, 
  author={Gangapurwala, S. and Geisert, M. and Orsolino, R. and Fallon, M. and Havoutis, I.},
  journal={IEEE Transactions on Robotics (T-RO)}, 
  year={2022},
  volume={38},
  number={5},
  pages={2908-2927},
  doi={10.1109/TRO.2022.3172469}
}

@inproceedings{nahrendra2023dreamwaq,
  title={Dreamwaq: Learning robust quadrupedal locomotion with implicit terrain imagination via deep reinforcement learning},
  author={Nahrendra, I. M. A. and Yu, B. and Myung, H.},
  booktitle={Proceedings of the IEEE International Conference on Robotics and Automation (ICRA)},
  pages={5078--5084},
  year={2023},
  organization={IEEE}
}

@inproceedings{kumar2021rma,
title={Rma: Rapid motor adaptation for legged robots},
author={Kumar, A. and Fu, Z. and Pathak, D. and Malik, J.},
booktitle={Proceedings of the Robotics: Science and Systems (RSS)},
year={2021}
}

@inproceedings{agarwal2023legged,
  title={Legged locomotion in challenging terrains using egocentric vision},
  author={Agarwal, A. and Kumar, A. and Malik, J. and Pathak, D.},
  booktitle={Proceedings of the Conference on Robot Learning (CoRL)},
  pages={403--415},
  year={2023},
  organization={PMLR}
}

@article{lee2020learning,
  title={Learning quadrupedal locomotion over challenging terrain},
  author={Lee, J. and Hwangbo, J. and Wellhausen, L. and Koltun, V. and Hutter, M.},
  journal={Science Robotics},
  volume={5},
  number={47},
  pages={eabc5986},
  year={2020},
  publisher={American Association for the Advancement of Science}
}

@INPROCEEDINGS{9981198,
  title={Advanced Skills by Learning Locomotion and Local Navigation End-to-End}, 
  author={Rudin, N. and Hoeller, D. and Bjelonic, M. and Hutter, M.},
  booktitle={Proceedings of the IEEE/RSJ International Conference on Intelligent Robots and Systems (IROS)}, 
  organization={IEEE},
  year={2022},
  pages={2497-2503},
  doi={10.1109/IROS47612.2022.9981198}
}

@inproceedings{rudin2022learning,
  title={Learning to walk in minutes using massively parallel deep reinforcement learning},
  author={Rudin, N. and Hoeller, D. and Reist, P. and Hutter, M.},
  booktitle={Conference on Robot Learning},
  pages={91--100},
  year={2022},
  organization={PMLR}
}

@article{cheng2023parkour,
title={Extreme Parkour with Legged Robots},
author={Cheng, X. and Shi, K. and Agarwal, A. and Pathak, D.},
journal={arXiv preprint arXiv:2309.14341},
year={2023}
}

@article{kim2021quadruped,
  title={Quadruped Locomotion on Non-Rigid Terrain using Reinforcement Learning},
  author={Kim, T. and Lee, S.},
  journal={arXiv preprint arXiv:2107.02955},
  year={2021}
}

@ARTICLE{8957061,
  title={STANCE: Locomotion Adaptation Over Soft Terrain}, 
  author={Fahmi, S. and Focchi, M. and Radulescu, A. and Fink, G. and Barasuol, V. and Semini, C.},
  journal={IEEE Transactions on Robotics (T-RO)}, 
  year={2020},
  volume={36},
  number={2},
  pages={443-457},
  doi={10.1109/TRO.2019.2954670}}

@inproceedings{yao2023predict,
  title={Predict the Physics-Informed Terrain Properties Over Deformable Soils using Sensorized Foot for Quadruped Robots},
  author={Yao, C. and Shi, G. and Ge, Y. and Zhu, Z. and Jia, Z.},
  booktitle={Proceedings of the International Conference on Advanced Robotics and Mechatronics},
  pages={330--335},
  year={2023},
  organization={IEEE}
}

@article{Suyoung2023deformter,
title = {Learning quadrupedal locomotion on deformable terrain},
author = {Choi, S.  and Ji, G.  and Park, J.  and Kim, H.  and Mun, J.  and Lee, J. and Hwangbo, J },
journal = {Science Robotics},
volume = {8},
number = {74},
pages = {eade2256},
year = {2023}
}

@ARTICLE{8772165,
  author={Jenelten, F. and Hwangbo, J. and Tresoldi, F. and Bellicoso, C. D. and Hutter, M.},
  journal={IEEE Robotics and Automation Letters (RA-L)}, 
  title={Dynamic Locomotion on Slippery Ground}, 
  year={2019},
  volume={4},
  number={4},
  pages={4170-4176},
  doi={10.1109/LRA.2019.2931284}}

@article{Focchi2018slip,
  title={Slip detection and recovery for quadruped robots},
  author={Focchi, M. and Barasuol, V. and Frigerio, M. and Caldwell, D. and Semini, C.},
  journal={Robotics Research:},
  volume={2},
  pages={185--199},
  year={2018},
  publisher={Springer}
}

@Article{s22082967,
AUTHOR = {Nisticò, Y. and Fahmi, S. and Pallottino, L. and Semini, C. and Fink, G.},
TITLE = {On Slip Detection for Quadruped Robots},
JOURNAL = {Sensors},
VOLUME = {22},
YEAR = {2022},
NUMBER = {8},
ARTICLE-NUMBER = {2967},
URL = {https://www.mdpi.com/1424-8220/22/8/2967},
PubMedID = {35458952},
ISSN = {1424-8220},
DOI = {10.3390/s22082967}
}

@article{iqbal2020provably,
  title={Provably stabilizing controllers for quadrupedal robot locomotion on dynamic rigid platforms},
  author={Iqbal, A. and Gao, Y. and Gu, Y.},
  journal={IEEE/ASME Transactions on Mechatronics},
  volume={25},
  number={4},
  pages={2035--2044},
  year={2020},
  publisher={IEEE}
}

@article{iqbal2021extended,
  title={Analytical solution to a time-varying LIP model for quadrupedal walking on a vertically oscillating surface},
  author={Iqbal, A. and Veer, S. and Gu, Y.},
  journal={Mechatronics},
  volume={96},
  pages={103073},
  year={2023},
  publisher={Elsevier}
}

@inproceedings{li2022posture,
  title={Posture Stabilization Control for a Quadruped Robot Walking on Swaying Platforms},
  author={Li, J. and Ye, L. and Jin, Z. and Liu, H. and Liang, B.},
  booktitle={Proceedings of the International Conference on Automation Science and Engineering (CASE)},
  pages={1959--1964},
  year={2022},
  organization={IEEE}
}

@article{grimminger2020open,
  title={An open torque-controlled modular robot architecture for legged locomotion research},
  author={Grimminger, F. and Meduri, A. and Khadiv, M. and Viereck, J. and Naveau, M. and others},
  journal={IEEE Robotics and Automation Letters (RA-L)},
  volume={5},
  number={2},
  pages={3650--3657},
  year={2020},
  publisher={IEEE}
}

@article{lee2021reinforcement,
  title={Reinforcement learning and neural network-based artificial intelligence control algorithm for self-balancing quadruped robot},
  author={Lee, C. and An, D.},
  journal={Journal of Mechanical Science and Technology},
  volume={35},
  pages={307--322},
  year={2021},
  publisher={Springer}
}

@article{sun2021adaptive,
  title={Adaptive quadruped balance control for dynamic environments using maximum-entropy reinforcement learning},
  author={Sun, H. and Fu, T. and Ling, Y. and He, C.},
  journal={Sensors},
  volume={21},
  number={17},
  pages={5907},
  year={2021},
  publisher={MDPI}
}

@inproceedings{zhuang2023robot,
  author    = {Zhuang, Z. and Fu, Z. and Wang, J. and Atkeson, C. and Schwertfeger, S. and Finn, C. and Zhao, H.},
  title     = {Robot Parkour Learning},
  booktitle = {Proceedings of the Conference on Robot Learning (CoRL)},
  year      = {2023},
}

@inproceedings{ma2023learning,
  title={Learning arm-assisted fall damage reduction and recovery for legged mobile manipulators},
  author={Ma, Y. and Farshidian, F. and Hutter, M.},
  booktitle={Proceedings of the IEEE International Conference on Robotics and Automation (ICRA)},
  pages={12149--12155},
  year={2023},
  organization={IEEE}
}

@inproceedings{yu2018policy,
title={Policy Transfer with Strategy Optimization},
author={Yu, W. and Liu, K. and Turk, G.},
booktitle={Proceedings of the International Conference on Learning Representations (ICLR)},
year={2019},
}

@inproceedings{peng2020learning,
  title={Learning agile robotic locomotion skills by imitating animals},
  author={Peng, X. B. and Coumans, E. and Zhang, T. and Lee, T-W and Tan, J. and Levine, S.},
  booktitle={Proceedings of the Robotics: Science and Systems (RSS)},
  year={2020}
}

@inproceedings{yu2019sim,
  title={Sim-to-real transfer for biped locomotion},
  author={Yu, W. and Kumar, V. C. and Turk, G. and Liu, C. K.},
  booktitle={Proceedings of the IEEE/RSJ International Conference on Intelligent Robots and Systems (IROS)},
  pages={3503--3510},
  year={2019},
  organization={IEEE}
}

@article{yu2020learning,
  title={Learning fast adaptation with meta strategy optimization},
  author={Yu, W. and Tan, J. and Bai, Y. and Coumans, E. and Ha, S.},
  journal={IEEE Robotics and Automation Letters (RA-L)},
  volume={5},
  number={2},
  pages={2950--2957},
  year={2020},
  publisher={IEEE}
}

@inproceedings{Yu2017si,
title={Preparing for the Unknown: Learning a Universal Policy with Online System Identification},
author={Yu, W. and Tan, J. and Liu, K. and Turk, G.},
booktitle={Proceedings of the Robotics: Science and Systems (RSS)},
year={2017},
}

@article{ji2022concurrent,
  title={Concurrent training of a control policy and a state estimator for dynamic and robust legged locomotion},
  author={Ji, G. and Mun, J. and Kim, H. and Hwangbo, J.},
  journal={IEEE Robotics and Automation Letters (RA-L)},
  volume={7},
  number={2},
  pages={4630--4637},
  year={2022},
  publisher={IEEE}
}

@inproceedings{fu2023deep,
  title={Deep whole-body control: learning a unified policy for manipulation and locomotion},
  author={Fu, Z. and Cheng, X. and Pathak, D.},
  booktitle={Proceedings of the Conference on Robot Learning (CoRL)},
  pages={138--149},
  year={2023},
  organization={PMLR}
}

@inproceedings{wang2023learning,
  title={Learning Robust, Agile, Natural Legged Locomotion Skills in the Wild},
  author={Wang, Y. and Jiang, Z. and Chen, J.},
  booktitle={RoboLetics: Workshop on Robot Learning in Athletics @ CoRL},
  year={2023}
}

@article{miki2022learning,
  title={Learning robust perceptive locomotion for quadrupedal robots in the wild},
  author={Miki, T. and Lee, J. and Hwangbo, J. and Wellhausen, L. and Koltun, V. and Hutter, M.},
  journal={Science Robotics},
  volume={7},
  number={62},
  pages={eabk2822},
  year={2022},
  publisher={American Association for the Advancement of Science}
}

@article{makoviychuk2021isaac,
  title={Isaac gym: High performance gpu-based physics simulation for robot learning},
  author={Makoviychuk, V. and Wawrzyniak, L. and Guo, Y. and Lu, M. and others},
  journal={arXiv preprint arXiv:2108.10470},
  year={2021}
}

@misc{wang2024unitree,
  author = {Wang, W.},
  title = {Unitree Robotics},
  howpublished = {\url{https://www.unitree.com/}},
  year = {2024},
  note = {Accessed on 5th January 2024}
}

@article{lesort2018state,
  title={State representation learning for control: An overview},
  author={Lesort, T. and D{\'\i}az-Rodr{\'\i}guez, N. and Goudou, J. and Filliat, D.},
  journal={Neural Networks},
  volume={108},
  pages={379--392},
  year={2018},
  publisher={Elsevier}
}

@article{schulman2017proximal,
  title={Proximal policy optimization algorithms},
  author={Schulman, J. and Wolski, F. and Dhariwal, P. and Radford, A. and Klimov, O.},
  journal={arXiv preprint arXiv:1707.06347},
  year={2017}
}

@article{borase2021review,
  title={A review of PID control, tuning methods and applications},
  author={Borase, R. P. and Maghade, D. K. and Sondkar, S. Y. and Pawar, S. N.},
  journal={International Journal of Dynamics and Control},
  volume={9},
  pages={818--827},
  year={2021},
  publisher={Springer}
}

@article{margolis2022rapid,
  title={Rapid locomotion via reinforcement learning},
  author={Margolis, G. B. and Yang, G. and Paigwar, K. and Chen, T. and Agrawal, P.},
  journal={Proceedings of the Robotics: Science and Systems (RSS)},
  year={2022}
}
